%% file: neurips_2024.tex
\documentclass{article}
\PassOptionsToPackage{numbers}{natbib}

\usepackage[final]{neurips_2024}

\usepackage[utf8]{inputenc} 
\usepackage[T1]{fontenc}    
\usepackage{url}            
\usepackage{booktabs}       
\usepackage{amsfonts}       
\usepackage{nicefrac}       
\usepackage{microtype}      

\usepackage[utf8]{inputenc} %
\usepackage[T1]{fontenc}    %
\usepackage[usenames,dvipsnames]{xcolor}
\usepackage[colorlinks=true, citecolor=BlueViolet, linkcolor=BlueViolet, urlcolor=blue]{hyperref}%
\usepackage{booktabs}       %
\usepackage{amsfonts}       %
\usepackage{nicefrac}       %
\usepackage{microtype}      %
\usepackage{graphicx}
\usepackage{tabularx}
\usepackage{multicol}
\usepackage{multirow}
\usepackage{tabu}
\usepackage{amsmath}
\usepackage{xcolor,colortbl}
\usepackage{xspace}
\usepackage{subcaption}
\usepackage{sidecap}
\usepackage{afterpage}
\usepackage{floatpag}
\usepackage{textcomp}
\usepackage{bm}

\definecolor{myred}{RGB}{255,0,0}
\definecolor{mygreen}{RGB}{112,173,71}
\newcommand{\red}[1]{{\color{red}#1}}
\definecolor{pink}{RGB}{225,20,147}

\makeatletter
\def\thanks#1{\protected@xdef\@thanks{\@thanks
        \protect\footnotetext{#1}}}

\title{3DGS-Enhancer: Enhancing Unbounded 3D Gaussian Splatting with View-consistent 2D Diffusion Priors}

\author{Xi Liu*~~~~~~~~~~~~~Chaoyi Zhou*~~~~~~~~~~~~~Siyu Huang\thanks{\emph{Corresponding author: Siyu Huang}}\\
  Visual Computing Division\\School of Computing\\
  Clemson University\\
  \texttt{\{xi9, chaoyiz, siyuh\}@clemson.edu} \\
}

\begin{document}

\maketitle

\newcommand{\teaserwidth}{0.25\textwidth}
\input{figures/teaser}

\begin{abstract}
 Novel-view synthesis aims to generate novel views of a scene from multiple input images or videos, and recent advancements like 3D Gaussian splatting (3DGS) have achieved notable success in producing photorealistic renderings with efficient pipelines. However, generating high-quality novel views under challenging settings, such as sparse input views, remains difficult due to insufficient information in under-sampled areas, often resulting in noticeable artifacts. This paper presents 3DGS-Enhancer, a novel pipeline for enhancing the representation quality of 3DGS representations. We leverage 2D video diffusion priors to address the challenging 3D view consistency problem, reformulating it as achieving temporal consistency within a video generation process. 3DGS-Enhancer restores view-consistent latent features of rendered novel views and integrates them with the input views through a spatial-temporal decoder. The enhanced views are then used to fine-tune the initial 3DGS model, significantly improving its rendering performance. Extensive experiments on large-scale datasets of unbounded scenes demonstrate that 3DGS-Enhancer yields superior reconstruction performance and high-fidelity rendering results compared to state-of-the-art methods. The project webpage is \href{https://xiliu8006.github.io/3DGS-Enhancer-project/}{\color{pink}https://xiliu8006.github.io/3DGS-Enhancer-project}.
\end{abstract}

\section{Introduction}
Novel-view synthesis (NVS) has decades of history in computer vision and graphics communities, aiming to generate views of a scene from multiple input images or videos. Recently, 3D Gaussian splatting (3DGS) \cite{kerbl3Dgaussians} has excelled in producing photorealistic renderings with a highly efficient rendering pipeline. However, rendering high-quality novel views far from existing viewpoints remains very challenging, as often encountered in sparse-view settings, due to insufficient information in under-sampled areas. As shown in Figure \ref{fig:teaser}, noticeable ellipsoid-like and hollow artifacts manifest when there are only three input views. Due to these common low-quality rendering results in practice, it is essential to enhance 3DGS to ensure its viability for real-world applications.

To our knowledge, few prior studies have specifically focused on enhancement methods aimed at improving the rendering quality of NVS.
Most existing enhancement work for NVS \cite{li2024dngaussian, Yang2023FreeNeRF} focuses on incorporating additional geometric constraints such as depth and normal into the 3D reconstruction process to fulfill the gap between the observed and unobserved regions. For example, DNGaussian \cite{li2024dngaussian} applies a hard-and-soft depth regularization to the geometry of radiance fields. However, these methods heavily rely on the effectiveness of additional constraint and are often sensitive to noises.
Another line of work leverages generative priors to regularize the  NVS pipeline. For instance, ReconFusion \cite{wu2024reconfusion} enhances Neural Radiance Fields (NeRFs) \cite{mildenhall2020nerf} by synthesising the geometry and texture for the unobserved regions. Although it can generate photo-realistic novel views, the view consistency is still challenging when the generated views are far away from the input ones.
  
In this work, we exploit the 2D generative priors, \textit{e.g.}, the latent diffusion models (LDMs) \cite{Rombach_2022_CVPR}, for 3DGS representation enhancement. LDM has demonstrated powerful and robust generation capabilities in various image generation \cite{Rombach_2022_CVPR} and restoration tasks \cite{xia2023diffir}. Nevertheless, the main challenge lies in the poor 3D view consistency among generated 2D images, which significantly hinders the 3DGS training process that requires highly precise view consistency. Although some efforts have been made, such as the Score Distillation Sampling (SDS) loss \cite{poole2023dreamfusion} that distills the optimization objective of a pre-trained diffusion model, it fails to generate the 3D representation allowing rendering high-fidelity images

Motivated by the analogy of the visual consistency between multi-view images and the temporal consistency between video frames, we propose to reformulate the challenging 3D consistency problem as an easier task of achieving temporal consistency within video generation, so we can leverage the powerful video diffusion models for restoring high-quality and view-consistent images. We propose a novel 3DGS enhancement pipeline, dubbed 3DGS-Enhancer. The core of 3DGS-Enhancer is a video LDM consisting of an image encoder that encodes latent features of rendered views, a video-based diffusion model that restores temporally consistent latent features, and a spatial-temporal decoder that effectively integrates the high-quality information in original rendered images with the restored latent features. The initial 3DGS model will be finetuned by these enhanced views to improve its rendering performance.
The proposed 3DGS-Enhancer can be trajectory-free to reconstruct the unbound scenes from sparse views and generate the natural 3D representation for the invisible area between two known views. 
A cocurrent work V3D \cite{chen2024v3d} also leverages latent video diffusion models \cite{blattmann2023stable} for generating object-level 3DGS models from single images. In contrast, our 3DGS-Enhancer focuses on enhancing any existing 3DGS models and thus can be applied to more generalized scenes, \textit{e.g.}, the unbounded outdoor scenes.

In experiments, we generate large-scale datasets with pairs of low-quality and high-quality images on hundreds of unbounded scenes, based on  DL3DV \cite{ling2023dl3dv}, for comprehensively evaluating the novelly investigated 3DGS enhancement problem. Empirical results demonstrate that the proposed 3DGS-Enhancer method achieves superior reconstruction performance on various challenging scenes, yielding more distinct and vivid rendering results. The code and the generated dataset will be publicly available. 
The contributions of this paper are summarized as follows.
\begin{enumerate}
    \item To the best of our knowledge, this is the first work to tackle the problem of enhancing low-quality 3DGS rendering results, an issue that widely exists in practical 3DGS applications. 
    \item We propose a novel pipeline 3DGS-Enhancer that addresses the 3DGS enhancement problem. 3DGS-Enhancer reformulates the 3D-consistent image restoration task as temporally consistent video generation, such that powerful video LDMs can be leveraged for generating both high-quality and 3D-consistent images. Novel 3DGS fine-tuning strategies are also devised for an effective integration of the enhanced views with the original 3DGS representation.
    \item We conduct extensive experiments on large-scale datasets of unbounded scenes to demonstrate the effectiveness of the proposed methods over existing state-of-the-art few-shot NVS methods.
\end{enumerate}

\section{Related Work}
\noindent\textbf{Radiance fields for novel view synthesis.}
Novel view synthesis (NVS) aims to generate unseen viewpoints from a set of input images and camera information. Radiance fields methods, like NeRFs \cite{mildenhall2020nerf}, encode 3D scenes as radiance fields and use volume rendering for novel views, achieving high-fidelity results but at the cost of lengthy training and inference times. Improvements such as Mip-NeRF \cite{barron2021mip,barron2022mipnerf360} enhance rendering quality through anti-aliasing, while others \cite{chen2022tensorf, fridovich2022plenoxels, yu2021plenoctrees, mubarik2023hardware} focus on speeding up the processes. Recently, 3D Gaussian splatting (3DGS) \cite{kerbl3Dgaussians} has emerged, offering competitive rendering quality and significantly higher efficiency by representing scenes as 3D Gaussian spheres and using a fast differentiable splatting pipeline \cite{article}. However, 3DGS still requires high-quality and numerous input views for optimal reconstruction, which is often impractical.

\noindent\textbf{Few-shot novel view synthesis.}
Leveraging additional information is essential for generating novel views from sparse input images. Various approaches incorporate different regularization techniques to prevent 3D geometry from overfitting to the training views. \cite{li2024dngaussian, wang2023sparsenerf, niemeyer2022regnerf, long2022sparseneus} introduce extra geometric information, such as depth maps or coarse mesh, to enhance the robustness and performance of 3D reconstruction from sparse views. \cite{charatan2024pixelsplat, SRF} leverage the learned priors from multi-view stereo datasets as general priors to improve performance in sparse view reconstruction tasks. FreeNeRF \cite{Yang2023FreeNeRF} integrates frequency and occlusion regularization during training to mitigate overfitting issues in few-shot neural rendering. Similarly, DietPixelNeRF \cite{Jain_2021_ICCV} employs a semantic view consistency loss to ensure that all views share consistent semantics, thereby alleviating overfitting. However, these methods are highly sensitive to the network's performance, where incorrect depth estimations or inaccurate mesh reconstructions can significantly degrade the final output. 

\noindent\textbf{Diffusion priors for novel view synthesis.}
Recently, utilizing diffusion models as priors for few-shot novel view synthesis has proven to be an effective approach. DreamFusion \cite{poole2023dreamfusion} employs Score Distillation Sampling (SDS) with a pre-trained diffusion model to guide 3D object generation from text prompts \cite{tang2023dreamgaussian, zeronvs, yi2023gaussiandreamer}. Some works \cite{liu2023zero, shi2023zero123plus, shi2024mvdream} embed 3D awareness into 2D diffusion models to generate multi-view images, though these methods typically require large datasets \cite{zhou2018stereo} and significant training resources \cite{Jain_2021_ICCV, niemeyer2022regnerf}. ReconFusion \cite{wu2024reconfusion} leavarge the 2D diffusion priors to recover a high-fidelity NeRF from sparse input views. More advanced approaches leverage video diffusion models \cite{blattmann2023stable, ho2022imagen, ho2022video, melaskyriazi2024im3d} for few-shot NVS. For instance, AnimateDiff \cite{guo2023animatediff} fine-tunes diffusion models with additional camera motions using LoRA \cite{hu2022lora}, while methods like SVD-MV \cite{blattmann2023stable}, V3D \cite{voleti2024sv3d} and IM-3D \cite{melaskyriazi2024im3d} propose camera-controlled video diffusion models for object-level 3D generation. In contrast, our approach offers greater generalizability for unbounded outdoor scenes.

\noindent\textbf{Radiance fields enhancement.}
Several existing studies focus on enhancing NeRFs by addressing the limited detail preservation issue caused by insufficient or low-quality input data. NeRF-SR \cite{wang2022nerf} and Refsr-nerf \cite{Huang_2023_CVPR}  use a super-resolution network to upscale the training view images, allowing novel views to be synthesized at higher resolutions with appropriate details. Alignerf \cite{Jiang_2023_CVPR} introduce optical-flow network to solve the misalignment problem to enhance the performance. Some other approaches incorporate 2D diffusion priors into 3D reconstructions. For instance, DiffusionNeRF \cite{wynn-2023-diffusionerf} leverages a diffusion model to learn gradients of logarithms of RGBD patch priors, serving as regularized geometry and color for a scene. Nerfbusters \cite{Warburg_2023_ICCV} use diffusion priors to remove ghostly artifacts in the 3D gaussians. Our work aim to addresses the radiance fields enhancement problem by proposing a novel framework 3DGS-Enhancer, achieving superior enhancement performance for low-quality unbounded 3DGS representations.

\section{Preliminary of 3D Gaussian Splatting} 
Here, we briefly review the formulation and rendering process of 3DGS \cite{kerbl3Dgaussians}. 3DGS represents a scene as a set of anisotropic 3D Gaussian spheres, allowing high-fidelity NVS with extremely low rendering latency. A 3D Gaussian sphere includes a center position $\mathbf{\mu} \in \mathbb{R}^3$, a scaling factor $\mathbf{s} \in \mathbb{R}^3$, and a rotation quaternion $\mathbf{q} \in \mathbb{R}^4$, such that the Gaussian distribution is
\begin{equation}
    G(x) = e^{-\frac{1}{2}(x-\mu)^T \Sigma^{-1} (x-\mu)},
    \label{eq:1}
\end{equation}
where $\Sigma = RSS^TR^T$, $S$ is the scaling matrix determined by $\mathbf{s}$ and $R$ is the rotation matrix determined by $\mathbf{q}$. To additionally model the view-dependent appearance, the Gaussian sphere also includes spherical harmonics (SH) coefficients $\mathcal{C} \in \mathbb{R}^k$, where k is the number of SH functions, and an $\alpha \in \mathbb{R}$ for opacity. The color and opacity are also calculated by the Gaussian distribution illustrated in Eq. \ref{eq:1}.

For rendering, all the 3D Gaussian spheres are projected onto the 2D camera planes via a differentiable Gaussian splatting pipeline \cite{article}. Given the viewing transform matrix $W$ and Jacobian matrix $J$ of the affine approximation of the projective transformation, the covariance matrix $\Sigma'$ in camera coordinates is calculated as
\begin{equation}
    \Sigma' = JW \Sigma W^TJ^T.
\end{equation}
The differentiable splatting method efficiently projects the 3D Gaussian spheres to 2D Gaussian distributions, ensuring fast $\alpha$-blending for rendering and color supervision. For each pixel, the color is rendered by $M$ Gaussian spheres that overlap with the pixel on the 2D camera planes, sorted in the depth distance as
\begin{equation}
    C = \sum_{i \in M}\mathcal{C}_i\alpha_i \prod_{j=1}^{i-1}(1 - \alpha_i).
\end{equation} 

\section{Method}
\subsection{3DGS-Enhancer: An Overview}
This work studies the 3DGS enhancement problem. More specifically, given a 3DGS model trained on a scene consisting of input views $\{ I^{\text{ref}}_1, I^{\text{ref}}_2, \ldots, I^{\text{ref}}_{N_{\text{ref}}}\}$ and corresponding camera poses $\{\bm{p}^{\text{ref}}_1, \bm{p}^{\text{ref}}_2, \ldots, \bm{p}^{\text{ref}}_{N_{\text{ref}}}\}$, the goal of this work is to enhance a set of low-quality novel views $\{ I_1, I_2, I_3, \ldots, I_{N_{new}}\}$ rendered by the 3DGS model. The enhanced images further fine-tune the 3DGS model to improve its reconstruction and rendering quality.

\begin{figure}
  \centering
  \includegraphics[width=1\textwidth]{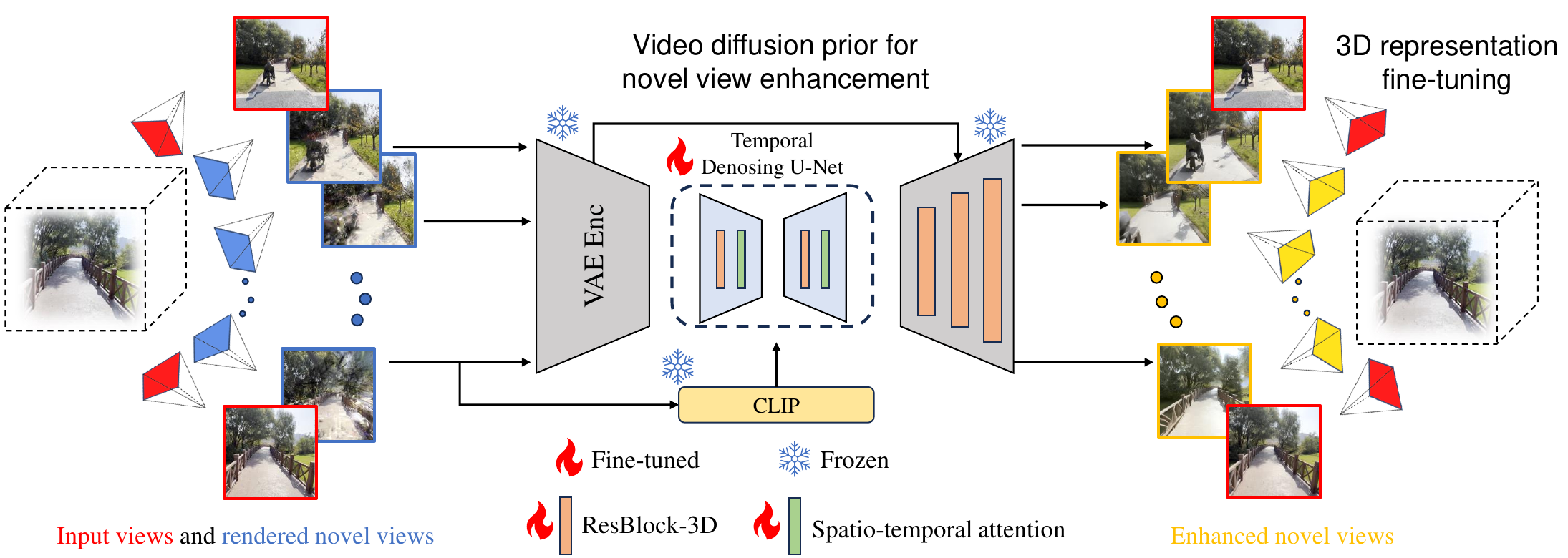}
  \caption{An overview of the proposed 3DGS-Enhancer framework for 3DGS representation enhancement. We learn 2D video diffusion priors on a large-scale novel view synthesis dataset to enhance the novel views rendered from the 3DGS model on a novel scene. Then, the enhanced views and input views jointly fine-tune the 3DGS model.}
  \label{fig:SVD}
\end{figure}

This work novelly reformulates the challenging task of 3D-consistent image restoration as the task of video restoration, in light of the analogy between the multi-view consistency and the video temporal consistency. We propose a novel framework named 3DGS-Enhancer that employs a video LDM comprising an image encoder, a video-based diffusion model, and a spatial-temporal decoder to enhance the rendered images while preserving a high 3D consistency. 3DGS-Enhancer also adopts novel fine-tuning strategies to selectively integrate the views enhanced by the video LDM into the 3DGS fine-tuning process. An illustration of the 3DGS-Enhancer framework is shown in Figure \ref{fig:SVD}. We discuss more details of the framework in the following.

\subsection{Video Diffusion Prior for Temporal Interpolation}
In this section, we introduce the video diffusion model for achieving 3D-consistent 2D image restoration. To lift the consistency between the generated 2D video frames and the high-quality reference views, we further propose to formulate the video restoration task as a video interpolation task, where the first frame and the last frame of inputs to the video diffusion model are two reference views. This formulation provides stronger guidance for the video restoration process.
Let $\{ \bm{p}^{\text{ref}}_{i-1}, \bm{p}^{s}_1, \bm{p}^{s}_2,\ldots, \bm{p}^{s}_T, \bm{p}^{\text{ref}}_i \}$ be the camera poses sampled from the trajectory fitted between two reference views, the images rendered accordingly are $v = \{ I^{\text{ref}}_{i-1}, I_1, I_2, \ldots, I_T, I^{\text{ref}}_i\}$. $v \in \mathbb{R}^{(T+2) \times 3 \times H \times W}$ serves as the input to the video diffusion model, \textit{e.g.}, a pre-trained image-guided stable video diffusion (SVD) model \cite{blattmann2023stable} that adopts cross-frame spatio-temporal attention module and 3D residual convolution in the diffusion U-Net. 
Unlike SVD, which repeats the single input image feature extracted by CLIP \cite{radford2021learning} for $T$ times as the conditional inputs, we input $\bm{v}$ to the CLIP encoder to get a sequence of conditional inputs $\bm{c}_{\text{clip}}$ and add it to the video diffusion model through cross attention. Meanwhile, we input $\bm{v}$ to the VAE encoder to get latent feature $\bm{c}_{\text{vae}}$ and add it into the diffusion model through a classifier-free guidance strategy to incorporate richer color information. The diffusion U-Net $\epsilon_{\theta}$ predicts the noise $\epsilon$ for each diffusion step $t$, and the training objective is 
\begin{equation}
    \mathcal{L}_{\text{diffusion}} = \mathbb{E}\left[ \| \epsilon - \epsilon_{\theta}(z_t, t, \bm{c}_{\text{clip}}, \bm{c}_{\text{vae}})\| \right].
\end{equation}
 where $z_t =\alpha_t z + \sigma_t \epsilon *..$ where z is the gt latent, $\epsilon \in \mathcal{N}(0, I)$, $\alpha_t$ and $\sigma_t$ define a noise at timestep $t$. The learned video diffusion model generates a sequence of enhanced image latents $z_v$ corresponding to the rendered low-quality views $v$. 

\subsection{Spatial-Temporal Decoder}
Although the video diffusion model can generate enhanced image latents $z_v$, we observe that there are artifacts such as temporal inconsistency, blurring, and color shift in outputs of the original decoder of video LDM. To address this issue, we propose a modified spatial-temporal decoder (STD). 
STD makes the following improvements over the original VAE Decoder: 
1) \textbf{Temporal decoding manner.} STD adopts additional temporal convolution layers to ensure the temporal consistency between decoded outputs. Similar to our video diffusion model, the first and the last input frames are the reference view images, and the intermediate inputs are the generated views; 
2) \textbf{Effective integration of rendered views.} STD adopts additional conditional inputs, same as those of the video diffusion model, allowing the decoder to better leverage the original rendered images. Inspired by \cite{yang2025motion, zhou2024upscale}, these conditional inputs are fed into STD through Controllable Feature Warping (CFW) modules \cite{wang2023exploiting}, such that their high-frequency patterns are better preserved.
3) \textbf{Color correction.} To address the color shift issue, we apply color normalization to the decoded images by following StableSR \cite{wang2023exploiting}. However, we observe that highly blurred and low-quality images in the conditional inputs can undermine the color correction effects. To mitigate this, we use the first reference view to calculate the mean and variance, and then align all the other decoded images with this reference view. Let $I^g_i$ be the $i$-th decoded images with a mean $\mathbf{u}_{\hat{I}^g_0}$ and a variance $\sigma_{\hat{I}^g_0}$, $\hat{I}^g_0$ be the reference view with a mean $\mu_{I^g_i}$ and a variance $\sigma_{I^g_i}$, the corrected image $I^c_i$ is computed by:
\begin{equation}
    I^c_i = \frac{I^g_i - \mu_{I^g_i}}{\sigma_{I^g_i}} \cdot  \sigma_{\hat{I}^g_0} +  \mathbf{u}_{\hat{I}^g_0}.
\end{equation}
The optimization objective of STD consists of an L1 reconstruction loss and an LPIPS perceptual loss between $I^g$ and ground-truth $\hat{I^g}$, and an adversarial loss, as
\begin{equation}
    \mathcal{L}_{\text{STD}} = \mathcal{L}_\text{rec} (I^g, \hat{I^g}) + \mathcal{L}_\text{LPIPS}(I^g, \hat{I^g}) + \mathcal{L}_\text{adv} (I^g). 
\end{equation}
where $\mathcal{L}_\text{adv}$ is the adversarial loss that discriminates between real image $\hat{I^g}$ and fake image $I^g$.

\subsection{Fine-tuning Strategies of 3D Gaussian Splatting}

\paragraph{Confidence-aware 3D Gaussian splatting.}    
Unlike existing sparse-view NVS methods, our approach does not rely on depth estimation networks for depth regularization. Instead, we take a purely 2D visual method by utilizing a video diffusion model to enhance images rendered from a low-quality 3DGS model. Despite this significant enhancement in the quality of the rendered views, we propose to rely more on the reference views rather than the restored novel views when fine-tuning the 3DGS model, since the 3DGS model is highly sensitive to slight inaccuracies in the restored views. These inaccuracies could be amplified during the fine-tuning process.

To minimize the negative impact of generated images on Gaussian training, we propose confidence-aware 3D Gaussian splatting. This strategy involves two levels of confidence, image level and pixel level. For the image level, the generated images that are closer to real images have lower confidence. For pixel level, the larger the mean covariance of all the Gaussians used to render this pixel, the higher its confidence. 

\paragraph{Image level confidence.}  In the task of novel view synthesis, if noise exists in two image views, a close distance between them increases the likelihood of generating conflicts and disrupting the 3D consistency of the scene. Therefore, for novel views that are close to the reference view, it is crucial to carefully optimize the 3D Gaussians to mitigate the adverse effects of noise. Conversely, when a novel view is far from all known views, it has a smaller likelihood of disturbing already well-reconstructed areas. Based on this reasoning, we normalize the distance from novel views to reference views between 0 and 1. The farther a viewpoint is from the reference view, the higher its confidence.

\begin{figure}
  \centering
  \includegraphics[width=1\textwidth]{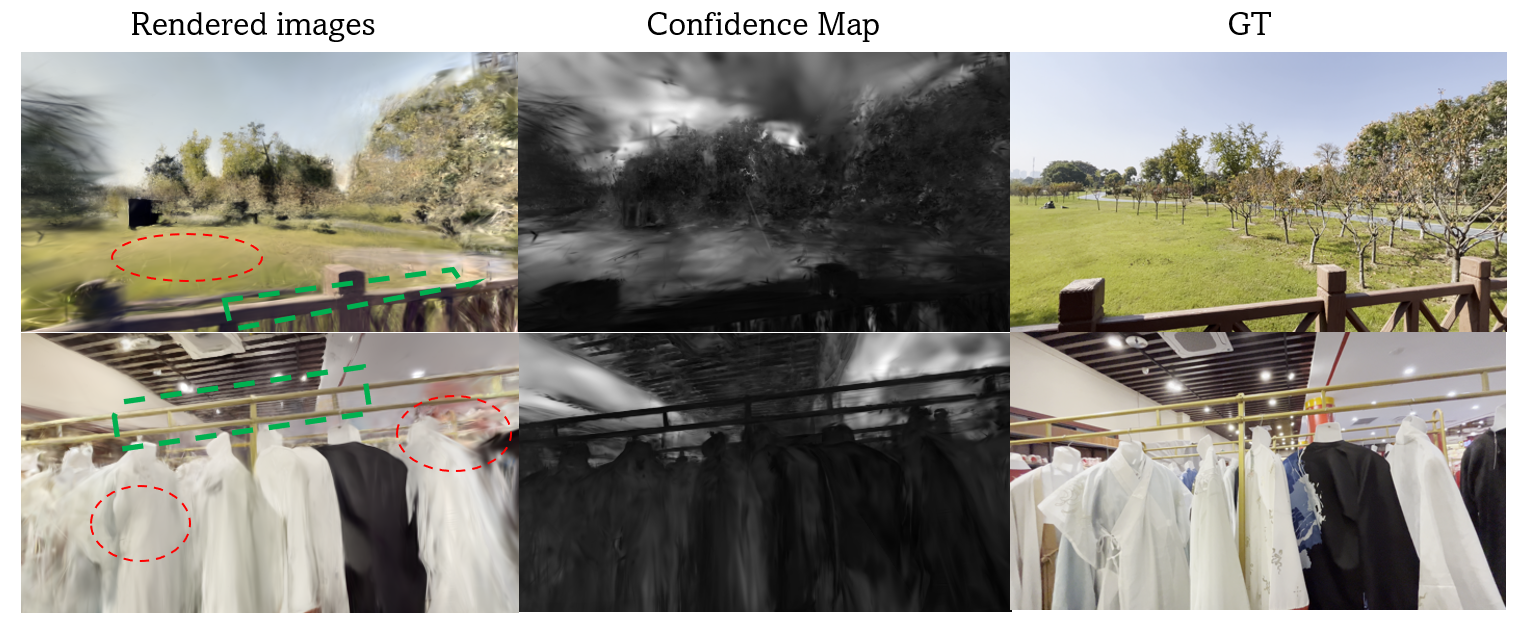}
  \caption{The \red{red circle} indicates the area with high confidence, meaning the generated videos can contribute more information. Conversely, the \textcolor{green}{ green quadrilateral } highlights the area with low confidence, suggesting that the generated video should not tend to optimize this area. }
  \label{fig: confidence_map}
\end{figure}

\paragraph{Pixel level confidence.} Inspired by ActiveNeRF \cite{pan2022activenerf}, which uses Gaussian distributions in NeRF to estimate uncertainty and identify views with the highest information gain, we aim to find the pixels that can provide the highest information gain from the generated images. As shown in Fig \ref{fig: confidence_map}, we observed that well-reconstructed areas are typically represented by Gaussians with very small volumes, calculated using the scaling vector $s \in \mathbb{R}$. Based on this observation, we propose a method to calculate pixel-level confidence.

The unique representation of 3D Gaussians allows us to render an H$\times$W$\times$3 image using a process similar to rendering colors, where each channel corresponds to one of the three components of the scaling vector $\mathbf{s}$. In 3DGS-Enhancer, we multiply these three channels of the scale map to obtain pixel-level confidence. For each pixel in the generated images, higher confidence results in greater weight in supervising the training of the 3DGS model.

Given a set of 3D Gaussian, the 3-channel $C_{conf}$ confidence map is rendered as same as colour rendering, and the formula is defined as follows
\begin{equation}
    C_{conf} = \sum_{i \in M}\mathbf{s}_i\alpha_i \prod_{j=1}^{i-1}(1 - \alpha_i).
\end{equation} 

And the 1 channel pixel level confidence map $ P_c = \sqrt[3]{C_{conf}[0] \times C_{conf}[1] \times C_{conf}[2]}$. Overall, in our training process for 3D Gaussians, the loss functions were defined as
\begin{equation}
     \mathcal{L}_{\text{3DGS}} = I_c \cdot ( P_c \odot \|C - \hat{C}\|_1 + SSIM(C, \hat{C}))
\end{equation}
where SSIM is the Structural Similarity Index and $\odot$ is Hadamard's product,  $I_c$ is the image-level confidence map and $\hat{C}$ is the real pixel value.



\begin{table*}[t]
\centering
\caption{\textbf{
A quantitative comparison of few-shot 3D reconstruction}. Experiments on DL3DV and LLFF follow the setting of \cite{Yang2023FreeNeRF}. Experiments on Mip-NeRF 360 follow the setting of \cite{wu2024reconfusion}. }
\vspace{-.5em}
\resizebox{1\textwidth}{!}{%
\huge
\begin{tabular}{l|ccc|ccc|ccc}
\toprule
 & \multicolumn{3}{c}{\textbf{3 views}} & \multicolumn{3}{|c|}{\textbf{6 views}} & \multicolumn{3}{c}{\textbf{9 views}} \\
 \textbf{Method} & PSNR$\uparrow$ & SSIM$\uparrow$ & LPIPS$\downarrow$ & PSNR$\uparrow$ & SSIM$\uparrow$ & LPIPS$\downarrow$ & PSNR$\uparrow$ & SSIM$\uparrow$ & LPIPS$\downarrow$ \\ \midrule

 \multicolumn{10}{c}{\textbf{DL3DV} (130 training scenes, 20 test scenes)} \\ \midrule 
 Mip-NeRF \cite{barron2021mip}     & 10.92 & 0.191 &  0.618  & 11.56 & 0.199 & 0.608  & 12.42 & 0.218 & 0.600  \\ 
 RegNeRF \cite{niemeyer2022regnerf}    & 11.46 & 0.214 & 0.600 & 12.69 & 0.236 & 0.579 & 12.33 &  0.219 &  0.598 \\
 FreeNeRF~ \cite{Yang2023FreeNeRF}         &  10.91 &  0.211 &  0.595  &  12.13 &  0.230 &  0.576 &  12.85 &  0.241 &  0.573    \\

 3DGS \cite{kerbl3Dgaussians}         & 10.97 &  0.248 &  0.567  &  13.34 &  0.332 &  0.498 & 14.99 &  0.403 &  0.446    \\
    
    DNGaussian \cite{li2024dngaussian} &  
   11.10 & 
   0.273 &
   0.579 &
   12.67 &  0.329 &
   0.547 &  13.44 &
   0.365 &  0.539    \\
 \textbf{3DGS-Enhancer (ours)} & \textbf{14.33} &\textbf{0.424 }&\textbf{0.464}  &\textbf{16.94}  &\textbf{ 0.565} &\textbf{0.356 } &\textbf{18.50}  &\textbf{0.630  }&\textbf{0.305} \\
 \bottomrule
\end{tabular}
}
\label{tab:main_table}
\end{table*}

\section{Experiments}
\subsection{3DGS-Enhance Dataset}
Given that the enhancement of 3DGS representations is a new task, we create a dataset to simulate various artifacts of the 3DGS representations. This dataset also serves as a more comprehensive benchmark for evaluating the performance of few-shot NVS methods. Existing few-shot NVS algorithms \cite{Yang2023FreeNeRF,li2024dngaussian} primarily focus on face-forward evaluations \cite{mildenhall2019llff}, where the test views have significant overlap with the input views. However, this evaluation method is not suitable for large-scale unbounded outdoor scenes. Therefore, we propose a dataset processing strategy that allows us to post-process any existing multi-view dataset to generate a large number of training image pairs that include typical artifacts caused by few-shot NVS. 

More specifically, for each scene, we have $n$ views $I_{\text{train}}=\{ I_1, I_2, \ldots, I_n\}$, which serve as the input for a high-quality 3DGS model. We uniformly sample a small number $m$ of views $I_{\text{low}}$ from $I_{\text{train}}$, which serve as the input for the low-quality 3DGS model. By linearly fitting the high-quality camera poses $p^{\text{train}}_i = \{ p^{\text{train}}_1, p^{\text{train}}_2, \ldots, p^{\text{train}}_{n^{*}}\}$, we randomly sample a camera trajectory $p^{\text{render}}_i = \{ p^{\text{render}}_1, p^{\text{render}}_2, \ldots, p^{\text{render}}_{n^{*}}\}$ on $p^{\text{train}}_i$ and render the image pairs using both high-quality and low-quality 3DGS models. This creates a set of high-quality and low-quality image pairs used for the training of our video diffusion model. 

We apply this dataset processing strategy to DL3DV \cite{ling2023dl3dv}, a large-scale outdoor dataset containing 10K scenes. We randomly select 130 scenes from the original DL3DV dataset and form more than 150,000 image pairs. We randomly select another 20 scenes from DL3DV  to form the test sets, evaluating the corss-scene capability of our method. More implementation details of the method can be found in the supplementary material.

\subsection{Comparison with State-of-the-Arts}

\input{figures/exp}
\input{rebuttal_fig_selected}

The quantitative and qualitative results on the DL3DV test set with 3 6 and 9 input views are shown in Table \ref{tab:main_table} and Figure \ref{fig:exp}.  
Our approach outperforms all the other baselines in PSNR, SSIM, and LPIPS scores. NeRF-based methods including Mip-NeRF \cite{barron2021mip} and FreeNeRF \cite{Yang2023FreeNeRF} produce blurry novel views due to smoothing inconsistencies. In contrast, 3DGS \cite{kerbl3Dgaussians} generates elongated elliptical artifacts due to local minima convergence. DNGuassian \cite{li2024dngaussian} reduces artifacts with depth regularization but results in blurry and noisy novel views. 

The first example in Figure \ref{fig:exp} demonstrates 3DGS-Enhacer's capability to remove artifacts while preserving view consistency. By interpolating input views using a video diffusion model, we incorporate more information while enrusing a high view consistency, enabling high-quality novel views and avoiding local minima. The second example highlights 3DGS-Enhancer's advantage in recovering high-frequency details. Our dataset processing strategy and video diffusion model enable an understand of strong multi-view prior across various scenes. As a result, very challenging cases such as the trees can be restored with sharp details. In summary, comparisons with baseline methods demonstrate our approach's potential to significantly improve the unbounded 3DGS representations, synthesizing high-fidelity novel views for open environments.

To demonstrate the generalizability of our method for out-of-distribution dataset, we train the methods on the DL3DV-10K dataset \cite{ling2023dl3dv} and test them on the Mip-NeRF360 dataset \cite{barron2022mipnerf360}. The results, as summarized in Table \ref{tab:comparison-Mip-NeRF360} and Fig \ref{fig:images-table-selected}, show that our method outperforms the baseline approaches, highlighting its remarkable generalization capabilities in unbounded environments.

\begin{table}[htbp]
\centering
\caption{A quantitative comparison of methods on the unseen Mip-NeRF360 dataset \cite{barron2022mipnerf360}. }
\resizebox{0.8\textwidth}{!}{%
\begin{tabular}{l|ccc|ccc}
\toprule
\multirow{2}{*}{\textbf{Method}} & \multicolumn{3}{c|}{\textbf{6 views}} & \multicolumn{3}{c}{\textbf{9 views}} \\
                                 & \textbf{PSNR} $\uparrow$ & \textbf{SSIM} $\uparrow$ & \textbf{LPIPS} $\downarrow$ & \textbf{PSNR} $\uparrow$ & \textbf{SSIM} $\uparrow$ & \textbf{LPIPS} $\downarrow$ \\
\midrule
    \multicolumn{7}{c}{\textbf{Mip-NeRF360} (all test scenes)} \\
\midrule
Mip-NeRF\                & 13.08                    & 0.159                     & 0.637                      & 13.73                    & 0.189                     & 0.628                      \\ 
RegNeRF                 & 12.69                    & 0.175                     & 0.660                      & 13.73                    & 0.193                     & 0.629                      \\ 
FreeNeRF                & 12.56                    & 0.182                     & 0.646                      & 13.20                    & 0.198                     & 0.635                      \\ 
3DGS                    & 11.53                    & 0.144                     & 0.651                      & 12.65                    & 0.187                     & 0.607                      \\ 
DNGaussian              & 11.81                    & 0.208                     & 0.689                      & 12.51                    & 0.228                     & 0.683                      \\
\textbf{3DGS-Enhancer (ours)}    & \textbf{13.96}           & \textbf{0.260}            & \textbf{0.570}             & \textbf{16.22}           & \textbf{0.399}            & \textbf{0.454}             \\
\bottomrule
\end{tabular}
}
\label{tab:comparison-Mip-NeRF360}
\end{table}

\subsection{Ablation Study}

\input{figures/ablation_table}

\paragraph{Real image as reference views.}
Table \ref{tab:ablation_table} shows the quantitative comparisons of different components in 3DGS-Enhancer framework. The video diffusion model provides strong multi-view priors. However, due to its native restrictions, we directly feed the original input views into the 3DGS fine-tuning process. This results in more reliable and view-consistent information from the input domain to facilitate 3DGS fine-tuning, as demonstrated by the "Real image" in Table \ref{tab:ablation_table}.

\paragraph{Confidence aware reweighting.}
Distant views are less likely to cause artifacts, so we normalize their distance to reference views between [0, 1], giving higher confidence of video diffusion results to farther viewpoints. This strategy is denoted by "Image confidence" in Table \ref{tab:ablation_table}. Pixel-level confidence, as denoted by "Pixel confidence" in Table \ref{tab:ablation_table}, is based on the density of small-volume Gaussians in well-reconstructed areas, using a color rendering pipeline to calculate volumes. Both pixel and image-level confidence strategies improve results individually, and their combination yields the best performance.

\input{figures/ablation_fig}

\paragraph{Video diffusion and STD.}
Figure \ref{fig:abation} visualizes the effects of video diffusion and STD module, respectively. Video diffusion removes most of the artifacts, and STD module enhances fine-grained and high-frequency textures, resulting in more vivid novel view renderings, which are closer to the ground truth. Table \ref{tab:ablation_std} shows the improvment for each modules.

\section{Conclusions, Limitations, and Future Work}
\label{sec:conclusion}
This paper has introduced 3DGS-Enhancer, a unified framework that applies view-consistency prior from video diffusion and use trajectory interpolation method to enhance unbounded 3DGS representations. By combining image and pixel-level confidence with 3DGS fine-tuning, we have achieved state-of-the-art performance in NVS enhancement. However, our approach relies on adjacent views for continuous interpolation, it cannot be easily adapted to single-view 3D model generation. 
Moreover, the confidence-aware 3DGS fine-tuning strategies are relatively simple and straightforward. In the future, it is interesting to integrate confidence maps directly with the video generation model, enabling the generation of images that are more in line with the real 3D world without the need for post-processing. Meanwhile, utilizing the efficient data generation capability of 3DGS to construct a massively scaled dataset for our video generation model presents a prime opportunity to enhance the model's 3D consistency. This approach also facilitates the 2D models to understand the 3D world directly from 2D images without additional geometric constraints. Regarding the social impact, the goal of this work is to advance the fields of 3D reconstruction and NVS. There are many potential societal consequences of our work, none which we feel must be specifically highlighted here.

\section{Acknowledgement}
The authors gratefully acknowledge the Clemson University Palmetto Cluster for providing the high-performance computing resources that supported the computations of this work.

\newpage
\bibliographystyle{plain}
\bibliography{neurips_2024}

\newpage
\section{Appendix}
    \input{appendix}
\end{document}

%% file: figures/teaser.tex
\newcommand{\smallwide}{1.0in}
\newcommand{\sevenwide}{1.6in}

\begin{figure*}[h!]
\footnotesize
\centering
\begin{tabular}{@{}c@{\,}c@{\,}c@{}}


\includegraphics[width=0.33\textwidth]{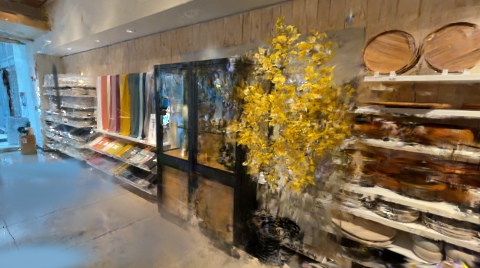} & 
\includegraphics[width=0.33\textwidth]{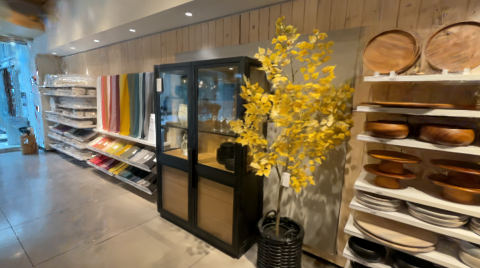} & 
\includegraphics[width=0.33\textwidth]{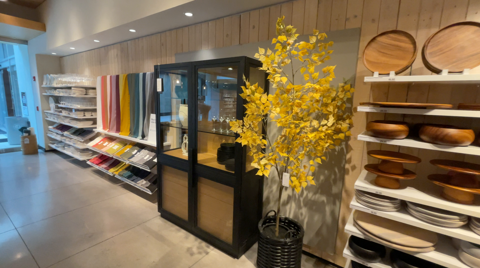} \\
\hspace{-0.3in} 3DGS (PSNR: 16.29) &  Ours (PSNR: 26.04) &  Ground-truth \vspace{.1em} \\

\includegraphics[width=0.33\textwidth]{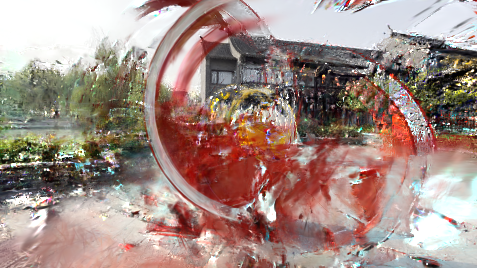} & 
\includegraphics[width=0.33\textwidth]{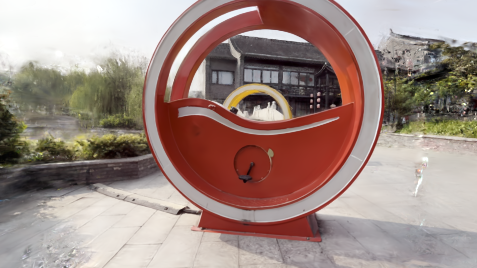} & 
\includegraphics[width=0.33\textwidth]{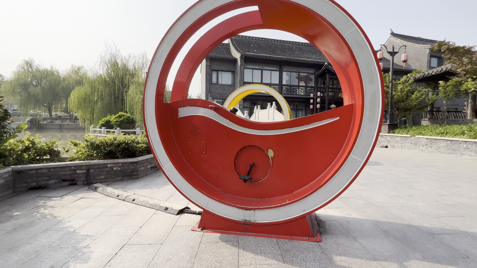} \\
\hspace{-0.3in} 3DGS (PSNR: 12.83) &  Ours (PSNR: 22.26) &  Ground-truth 

\end{tabular}
\caption{The 3DGS-Enhancer improves 3D Gaussian splatting representations on unbounded scenes with sparse input views.}
\label{fig:teaser}
\end{figure*}

%% file: figures/exp.tex
\begin{figure*}[t]
\footnotesize
\centering
\begin{tabular}{@{}c@{\,}c@{\,}c@{}}
\includegraphics[width=0.33\textwidth]{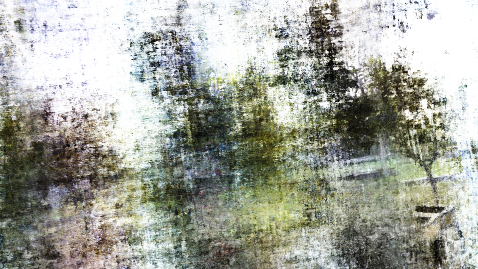} & 
\includegraphics[width=0.33\textwidth]{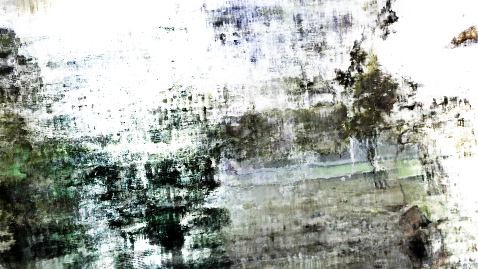} & 
\includegraphics[width=0.33\textwidth]{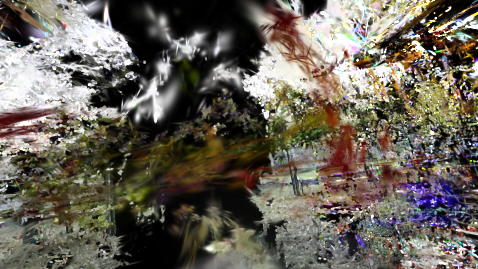} \\
\vspace{-1.4em} \\
\hspace{-0.3in} MipNerf \cite{barron2021mip} &  FreeNerf \cite{Yang2023FreeNeRF}&  3DGS \cite{kerbl3Dgaussians} \\
\includegraphics[width=0.33\textwidth]{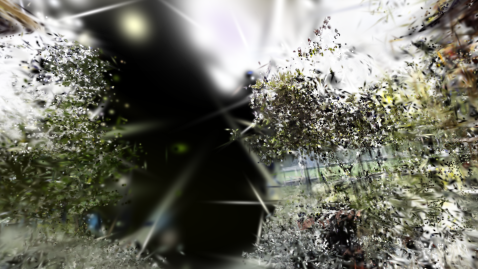} & 
\includegraphics[width=0.33\textwidth]{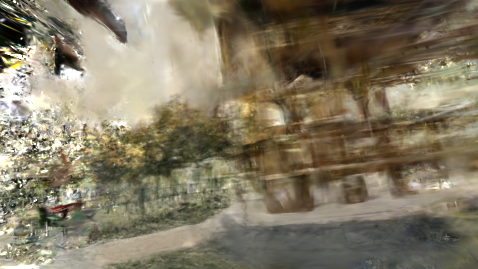} & 
\includegraphics[width=0.33\textwidth]{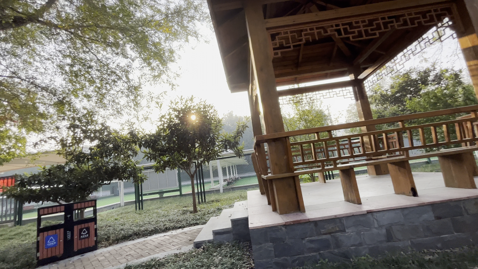} \\
\vspace{-1.4em} \\
\hspace{-0.3in} DNGaussian \cite{li2024dngaussian} &  Ours &  Ground-truth \\

\includegraphics[width=0.33\textwidth]{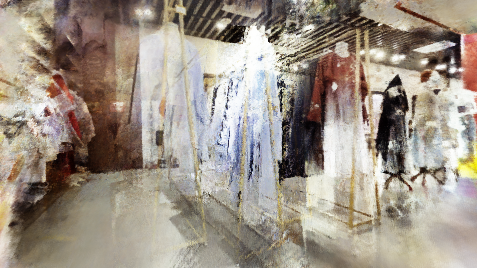} & 
\includegraphics[width=0.33\textwidth]{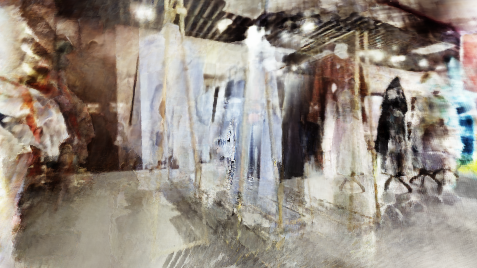} & 
\includegraphics[width=0.33\textwidth]{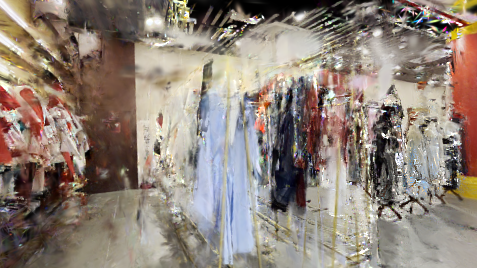} \\
\vspace{-1.4em} \\
\hspace{-0.3in} MipNerf \cite{barron2021mip} &  FreeNerf \cite{Yang2023FreeNeRF}&  3DGS \cite{kerbl3Dgaussians} \\
\includegraphics[width=0.33\textwidth]{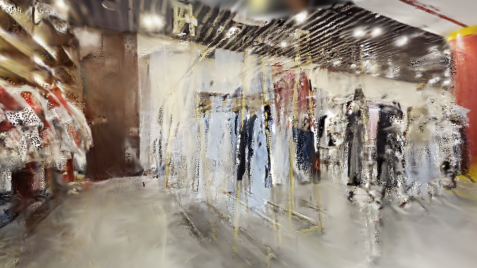} & 
\includegraphics[width=0.33\textwidth]{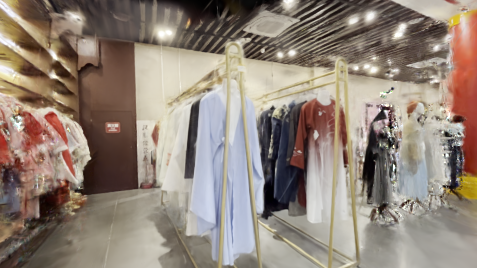} & 
\includegraphics[width=0.33\textwidth]{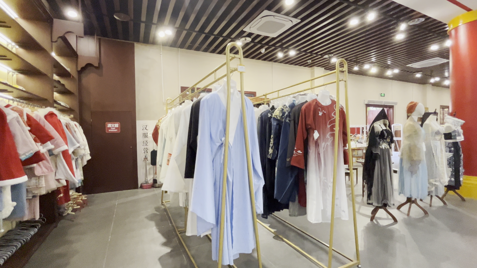} \\
\vspace{-1.4em} \\
\hspace{-0.3in} DNGaussian \cite{li2024dngaussian} &  Ours &  Ground-truth \\

\vspace{-1em} \\

\includegraphics[width=0.33\textwidth]{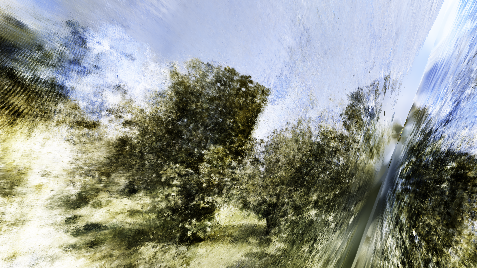} & 
\includegraphics[width=0.33\textwidth]{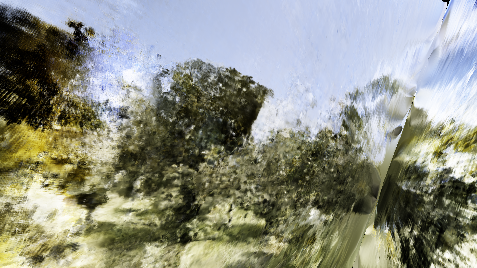} & 
\includegraphics[width=0.33\textwidth]{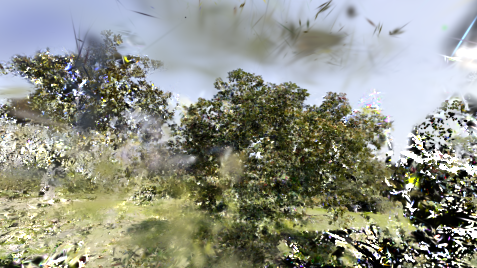} \\
\vspace{-1.4em} \\
\hspace{-0.3in} MipNerf \cite{barron2021mip} &  FreeNerf \cite{Yang2023FreeNeRF}&  3DGS \cite{kerbl3Dgaussians} \\
\includegraphics[width=0.33\textwidth]{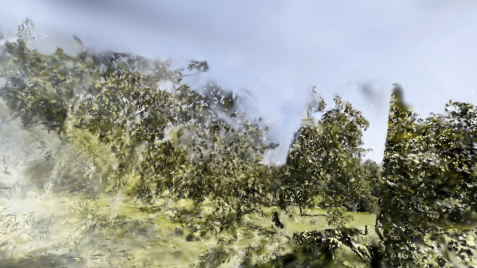} & 
\includegraphics[width=0.33\textwidth]{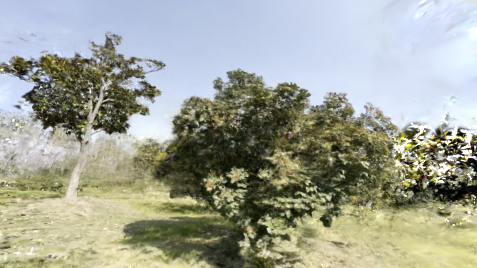} & 
\includegraphics[width=0.33\textwidth]{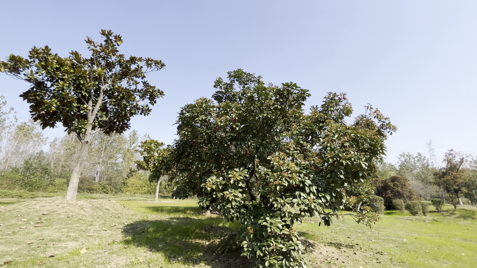} \\
\vspace{-1.4em} \\
\hspace{-0.3in} DNGaussian \cite{li2024dngaussian} &  Ours &  Ground-truth \\

\end{tabular}
\caption{A visual comparison of rendered images on scenes from DL3DV \cite{ling2023dl3dv} test set with the 3-view setting. }
\label{fig:exp}
\end{figure*}

%% file: rebuttal_fig_selected.tex
\begin{figure}[ht]
    \centering
    \captionsetup{type=figure}
    
    \begin{tabular}{c@{\hskip 0.1in}c@{\hskip 0.1in}c}
        \includegraphics[width=0.3\textwidth]{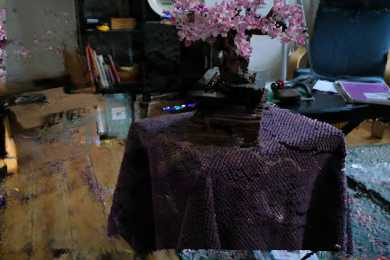} &  
        \includegraphics[width=0.3\textwidth]{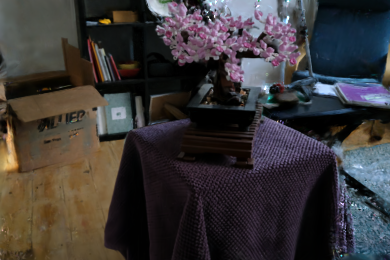} & 
        \includegraphics[width=0.3\textwidth]{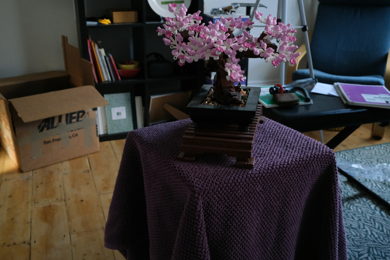} \\
        \small 3DGS (PSNR: 16.15)  & \small Ours (PSNR: 22.32) & \small Ground-truth \vspace{.5em} \\
    \end{tabular}
    
    \begin{tabular}{c@{\hskip 0.1in}c@{\hskip 0.1in}c}
        \includegraphics[width=0.3\textwidth]{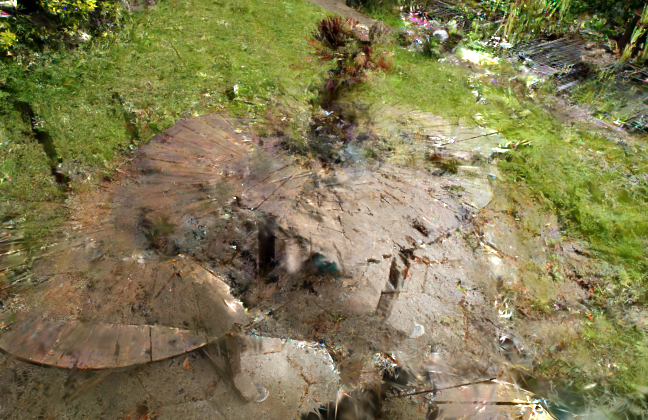} &  
        \includegraphics[width=0.3\textwidth]{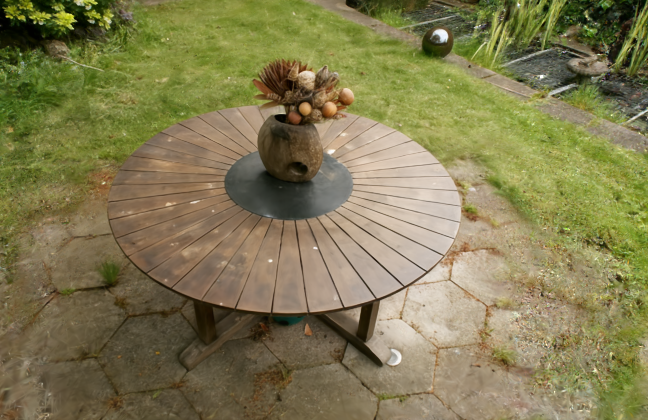} & 
        \includegraphics[width=0.3\textwidth]{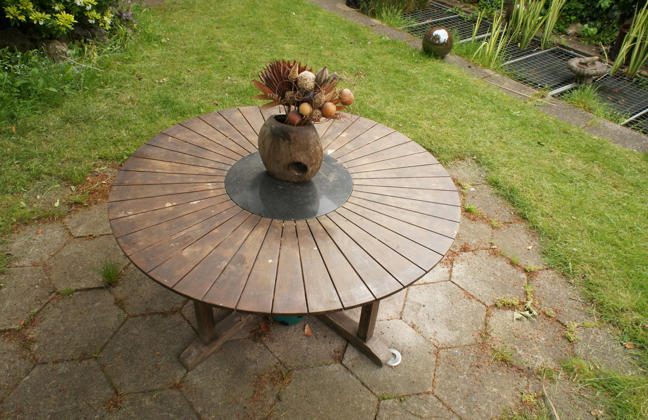} \\
        \small 3DGS (PSNR: 13.90)  & \small Ours (PSNR: 22.32) & \small Ground-truth \vspace{.5em} \\
    \end{tabular}
    
    \caption{A visual comparison of cross-dataset generalization ability, where the methods are trained on the DL3DV-10K dataset \cite{ling2023dl3dv} and tested on the Mip-NeRF360 dataset \cite{barron2022mipnerf360}.}
    \label{fig:images-table-selected}
\end{figure}

%% file: figures/ablation_table.tex
\begin{table*}[t]
\centering
\caption{ An ablation study of the four modules of our 3DGS-Enhancer framework, where all results are averaged across 3, 6, 9, and 12 input views on DL3DV dataset \cite{ling2023dl3dv}.}
\vspace{-1em}
\resizebox{1\textwidth}{!}{%
\begin{tabular}{cccccc|ccccc}
\cmidrule[\heavyrulewidth]{2-9}
  && Video diffusion  & Real image & Image confidence  & Pixel confidence    & PSNR$\uparrow$ & SSIM$\uparrow$ & LPIPS$\downarrow$  \\ \cmidrule{2-9}
   &   & \checkmark&  - & - & - &14.33   &0.476  &0.422  &   \\
   &   & \checkmark& \checkmark  & - & - &17.01   &0.553  &0.361   &   \\
   &  &\checkmark &  \checkmark & \checkmark & - &17.29   &0.570  &0.354   &   \\
   &  &\checkmark &  \checkmark &- &\checkmark  &  17.16 &0.564  & 0.351  &   \\
   &   &\checkmark &  \checkmark &\checkmark  &\checkmark  &  \textbf{17.34} & \textbf{0.574} & \textbf{0.351}  &   \\
   \cmidrule{2-9}

\end{tabular}
}
\label{tab:ablation_table}
\end{table*}

\begin{table}[htbp]
\centering
\caption{An ablation study of STD (temporal layers) and color correction module on the DL3DV test dataset with a 9-view setting.}
\resizebox{0.86\textwidth}{!}{%
\begin{tabular}{ccc|ccc}
\toprule
Video diffusion & STD (temporal layers) & color correction & PSNR $\uparrow$ & SSIM $\uparrow$ & LPIPS $\downarrow$ \\ 
\midrule
$\checkmark$  & -  & -  & 18.11  & 0.591  & 0.312  \\ 
$\checkmark$  & $\checkmark$  & -  & 18.44  & 0.625  & 0.306  \\ 
$\checkmark$  & $\checkmark$  & $\checkmark$  & \textbf{18.50}  & \textbf{0.630}  & \textbf{0.305}  \\ 
\bottomrule
\end{tabular}
}
\label{tab:ablation_std}
\end{table}

%% file: figures/ablation_fig.tex
\newcommand{\ablationwide}{1.2in}
\begin{figure*}[!t]
\vspace{1.5em}
\footnotesize
\centering
\begin{tabular}{cccc}
\includegraphics[width=\ablationwide]{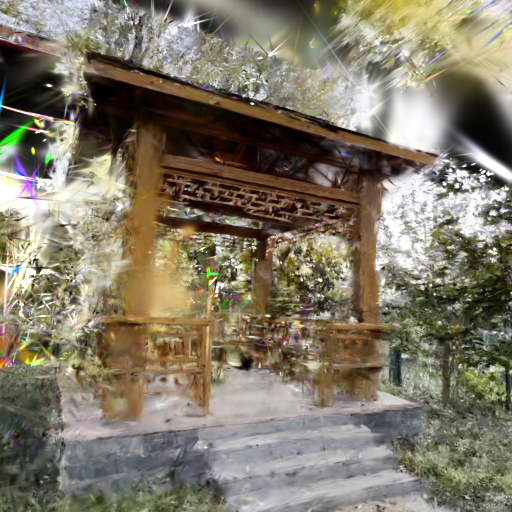}&
\includegraphics[width=\ablationwide]{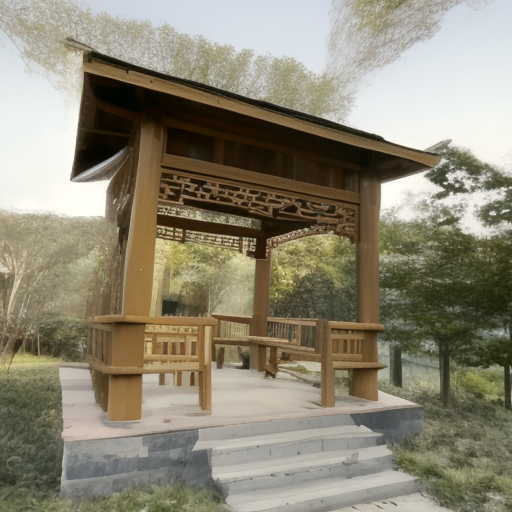} & 
\includegraphics[width=\ablationwide]{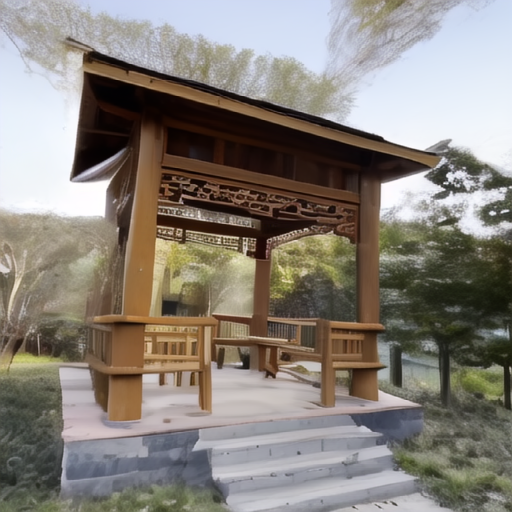} &\includegraphics[width=\ablationwide]{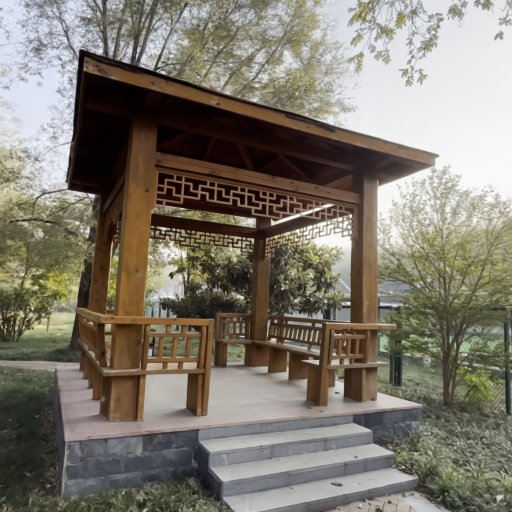}  \\
Input &  Diffusion &  STD & Ground-truth\\
\end{tabular}
\caption{An ablation study of the video diffusion model components in our 3DGS-Enhancer framework.}
\label{fig:abation}
\end{figure*}

%% file: appendix.tex
\subsection{Details of 3DGS Enhancement Dataset}
For our 3DGS Enhancement Dataset, constructed based on DL3DV, we randomly select 120 scenes to create the training set for our video diffusion model and 30 scenes as the test set. By following previous works, we use the standard train/test split, selecting every 8th frame of the remaining frames for evaluation.

To create image pairs simulating the artifacts due to the lack of input views in novel view synthesis problem, we render the image pairs from pairs of low-high quality 3DGS models. Specifically, the input views for the high-quality model consist of all images in the original dataset, while the inputs for the low-quality model are a subset uniformly sampled from the original dataset. To add more complexity, we sample the subset according to a certain number (e.g., 3, 6, 9) or a certain ratio (e.g., 5\%).
With the aim to fully capture the distribution of artifacts created by the sparse input views and train the video diffusion model with smoother inputs, we propose a heuristic trajectory fitting algorithm, as shown in Figure \ref{fig:trajectory}, proving a sequence of cameras by interpolating the low or high-quality model's input views. Specifically, if the original camera trajectories are smooth and simple, such as those of DL3DV, we use the high-quality input views as the reference to fit the trajectories. For complex trajectories, such as those in Mip-NeRF 360, we use the low-quality input to avoid significantly poor rendering results, which would lead to unreasonable artifact distributions.
As a result, we render a large number of image pairs with and without artifacts, as shown in Figure \ref{fig:pairs}, at a resolution of 512 × 512, leading to powerful video diffusion priors with high view consistency and photo-realism.

\input{figures/trajectory}

\input{figures/pairs}

\subsubsection{Training details}

Our video diffusion model includes a pre-trained VAE to encode an image sequence into a latent sequence and decode the latent sequence back into the image sequence. It also includes a U-Net with learnable temporal layers, which employs cross-frame attention modules and 3D CNN modules to ensure frame-consistent outputs. The input of video diffusion model is a image sequence segment that includes 25 images with different sample steps from the image sequences rendered from the low-quality 3DGS model. The first and the last frames in this segment are replaced with images rendered from the high-quality 3DGS model. During fine-tuning, our video diffusion model is conditioned on these image sequence segments and trained to synthesize the corresponding segments rendered from the high-quality 3DGS model.

Our video diffusion model is fine-tuned with a learning rate of 0.0001, incorporating 500 steps for warm-up, followed by a total of 80,000 training steps. The batch size is set to 1 in each GPU, where each batch consisted of 25 images at 512x512 resolution. To optimize the training process, the Adam optimizer is employed. Additionally, a dropout rate of 0.1 is applied to the conditions between the first and last frames and the training process utilize CFG (classifier-free guidance) to train the diffusion model. The training is conducted on 2 NVIDIA A100-80G GPUs over 3 days. The STD is fine-tuned with a learning rate of 0.0005 and 50,000 training steps. The batch size is set to 1 in each GPU, where each batch consists of 5 images at 512x512 resolution, but for inference, it was increased to 25. The fine-tuning process is conducted on 2 NVIDIA A100-80G GPUs in 2 days. The entire pipeline's inference and training speeds were evaluated and are presented in Table \ref{tab:runtime_fps}.

\begin{table}[htbp]
\centering
\caption{A comparison of per-scene training time and rendering FPS between methods. For our method, the LQ-3DGS reconstruction takes 10.5 minutes, stable video diffusion inference for 50 novel views requires 2.0 minutes, and the HQ-3DGS reconstruction takes 12.0 minutes.}
\begin{tabular}{c|c|c}
\toprule
\textbf{Method} & \textbf{Per-scene training time} $\downarrow$ & \textbf{Rendering FPS} $\uparrow$ \\ 

\midrule
Mip-NeRF        & 10.7h          & 0.09 \\ 
RegNeRF          & 2.5h           & 0.09 \\ 
FreeNeRF         & 3.8h           & 0.09 \\ 
3DGS             & 10.5min        & 100  \\ 
DNGaussian       & 3.3min         & 100  \\ 
3DGS-Enhancer (ours) & 24.5min  & 100 \\
\bottomrule
\end{tabular}
\label{tab:runtime_fps}
\end{table}

\subsection{Details of Comparison Baselines}
For the evaluation datasets, we compare against
the standard 3D Gaussian Splatting \cite{kerbl3Dgaussians}
(which is also the reconstruction pipeline used in our
work), and the state-of-the-art few-view NVS regularization
methods, including Mip-NeRF \cite{barron2021mip}, FreeNeRF \cite{Yang2023FreeNeRF}, Zip-NeRF \cite{barron2023zipnerf}, and
RegNeRF \cite{niemeyer2022regnerf}. We also compare to some few-shot NVS methods using generative priors including ZeroNVS \cite{zeronvs}, and ReconFusion \cite{wu2024reconfusion}.

For the evaluation of MipNeRF, FreeNeRF, RegNeRF, and DNGaussian  on DL3DV and  Mip-NeRF 360 dataset,  we follow the original configurations and code shared by the authors. Additionally,  we use random point cloud as the initialization for 3DGS, following the implementations from DNGaussian. We also decrease the batch size for RegNeRF from 4096 to 512 according to the limited computation resource. 

%% file: figures/trajectory.tex
\begin{figure*}[h]
\footnotesize
\centering
\begin{tabular}{@{}c@{\,}c@{\,}c@{}}
\includegraphics[width=\sevenwide]{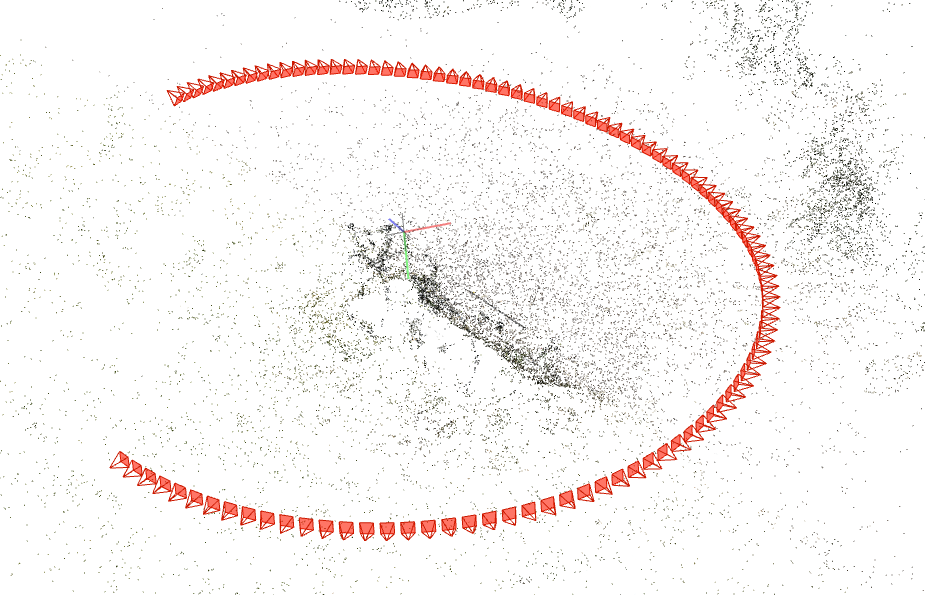} & 
\includegraphics[width=\sevenwide]{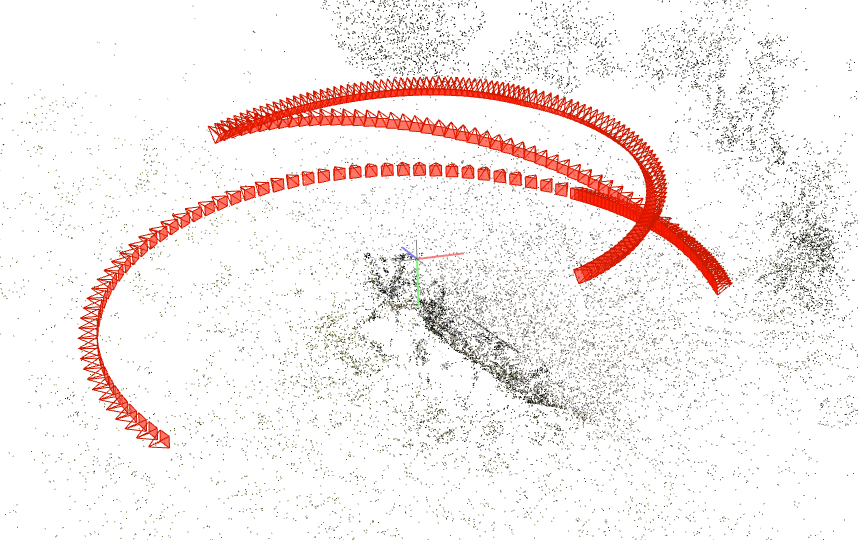} & 
\includegraphics[width=\sevenwide]{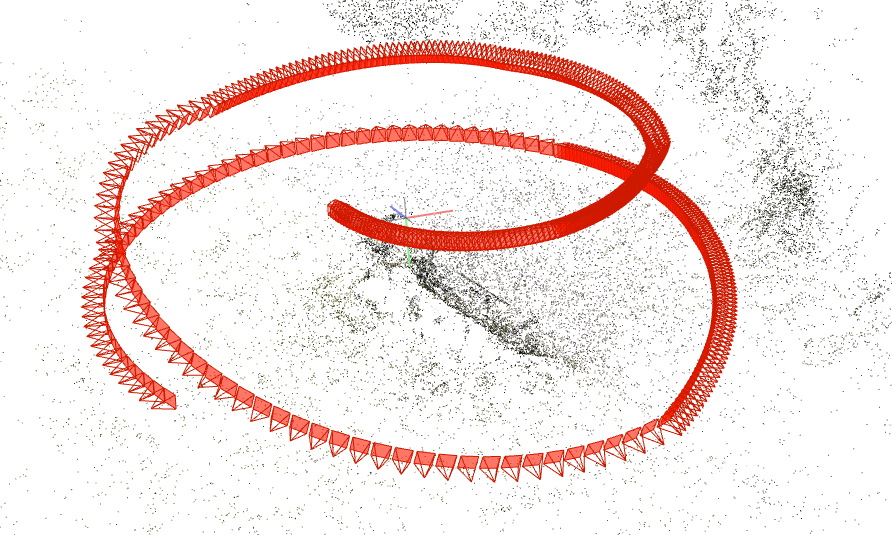} \\
\hspace{-0.3in} Input: 3 & Input: 6& Input: 9 \\
\end{tabular}
\caption{The fitting trajectories under different number of input views. }
\label{fig:trajectory}
\end{figure*}

%% file: figures/pairs.tex
\begin{figure*}[t]
\footnotesize
\centering
\begin{tabular}{@{}c@{\,}c@{\,}c@{\,}c@{}}
\hspace{-0.5in}
\includegraphics[width=\sevenwide]{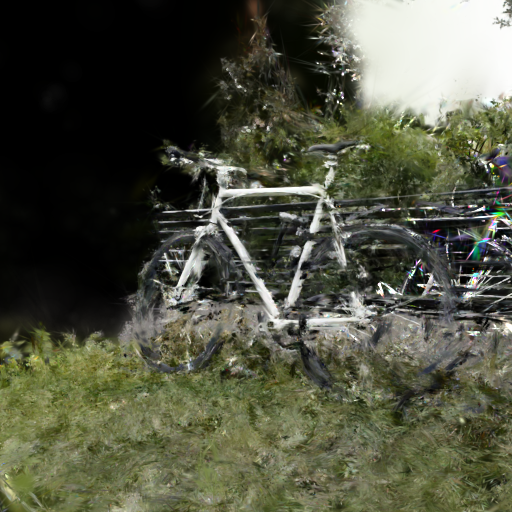} & 
\includegraphics[width=\sevenwide]{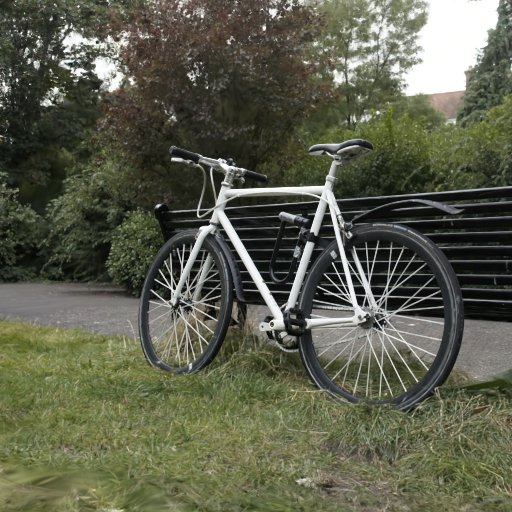} & 
\includegraphics[width=\sevenwide]{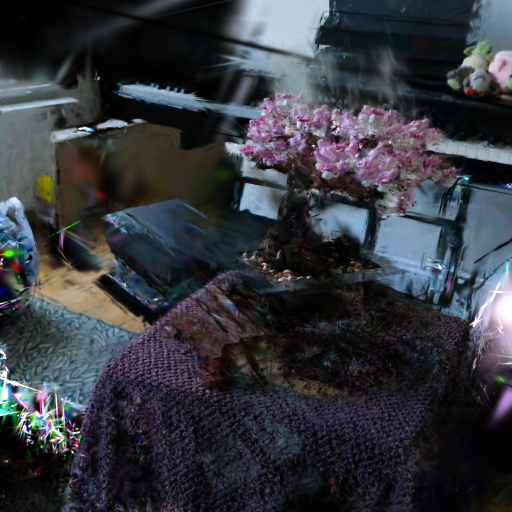} &
\includegraphics[width=\sevenwide]{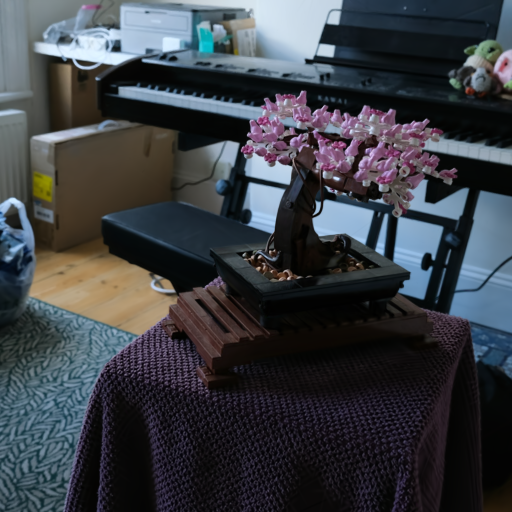}  \\
\hspace{-0.3in}Input: 3  &  Ground-truth &  Input: 6&  Ground-truth \\
\hspace{-0.5in}
\includegraphics[width=\sevenwide]{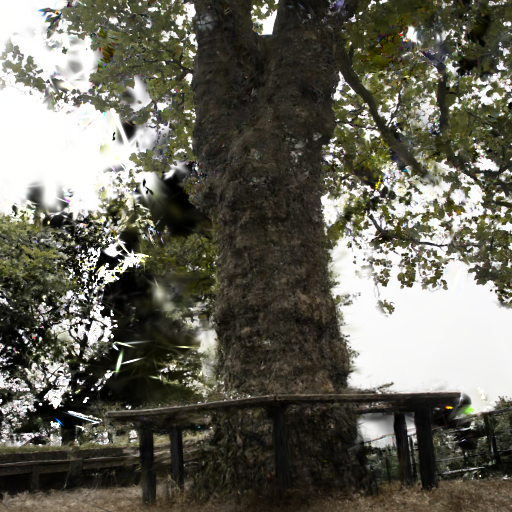} & 
\includegraphics[width=\sevenwide]{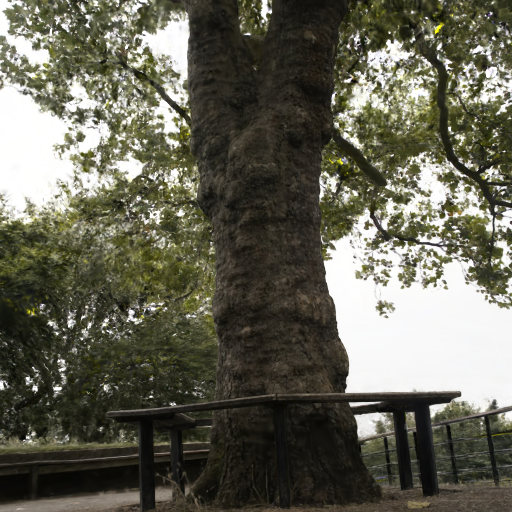} & 
\includegraphics[width=\sevenwide]{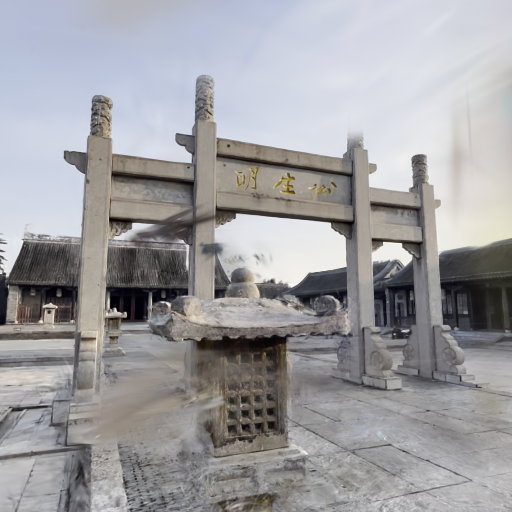} &
\includegraphics[width=\sevenwide]{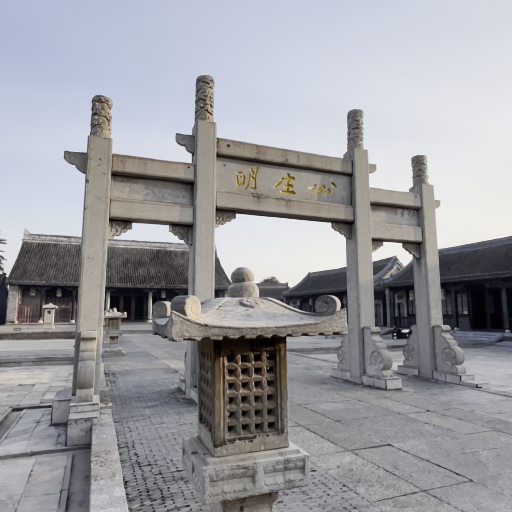}  \\
\hspace{-0.3in}Input: 9  &  Ground-truth &  Input: 5\%&  Ground-truth \\

\end{tabular}
\caption{The low and high quality image pairs created in our 3DGS Enhancement dataset. }
\label{fig:pairs}
\end{figure*}